\pdfoutput=1
\PassOptionsToPackage{table}{xcolor}
\documentclass[11pt]{article}

\usepackage{ACL2023}

\usepackage{times}
\usepackage{latexsym}

\usepackage[T1]{fontenc}

\usepackage[utf8]{inputenc}

\usepackage{microtype}

\usepackage{inconsolata}

\RequirePackage[inline,shortlabels]{enumitem}
\setlist{nosep}

\usepackage{amsmath, amssymb}
\usepackage{graphicx}
\usepackage{booktabs}
\usepackage{multicol, multirow}
\usepackage[ruled,linesnumbered]{algorithm2e}
\usepackage{bbm}
\usepackage{float}
\usepackage{caption}
\usepackage{xcolor}
\usepackage{natbib}
\usepackage{defs}
\usepackage{todonotes}

\SetCommentSty{mycommfont}
\SetKwComment{Comment}{/* }{ */}
\SetKwFor{For}{for (}{) $\lbrace$}{$\rbrace$}
\RestyleAlgo{ruled}

\newcommand{\sysname}{CoCoNTs} 
\newcommand{\NT}{NT}
\newcommand{\AllNTL}{AllNTs}
\newcommand{\dataset}[1]{\mathcal{D}_\text{#1}}
\newcommand{\trainset}{\dataset{train}}

\newcommand{\token}[1]{t_{#1}}
\newcommand{\prefix}[1]{[\token{1}\dots\token{#1}]}
\newcommand{\vocab}{\mathcal{V}}

\newcommand{\x}[1]{\mathbf{x}_{#1}}
\newcommand{\y}[1]{\mathbf{y}_{#1}}

\newcommand{\topr}[2]{\textrm{Top-}#1(#2)}

\newcommand{\vt}{\mathbf{t}}

\newcommand{\ymulti}[1]{\mathbf{y}^\text{all}_{#1}}
\newcommand{\yadj}[1]{\mathbf{y}^\text{CC}_{#1}}
\newcommand{\ytopr}{\vy^{\text{topr}}}
\newcommand{\ohot}[1]{\mathbf{1}_\vocab(#1)}
\newcommand{\seqx}[1]{[\x{1}\dots\x{#1}]}
\newcommand{\seqy}[1]{[\y{1}\dots\y{#1}]}

\title{Efficient Training of Language Models with Compact and Consistent Next Token Distributions}
\newcommand{\aspace}{\hspace{1.0em}}

\author{
Ashutosh Sathe \aspace Sunita Sarawagi\\
Indian Institute of Technology, Bombay \\
\texttt{\{absathe,sunita\}@cse.iitb.ac.in} \\
\url{https://github.com/ashutoshbsathe/CoCoNTs}
}

\begin{document}
\maketitle
\begin{abstract}
Maximizing the likelihood of the next token is an established, statistically sound objective for pre-training language models. 
In this paper we show that we can train better models faster by pre-aggregating the corpus with a collapsed n-gram distribution.
Previous studies have proposed corpus-level $n$-gram statistics as a regularizer; however, the construction and querying of such $n$-grams, if done naively, prove to be costly and significantly impede training speed, thereby limiting their application in modern large language model pre-training.

We introduce an alternative compact representation of the next token distribution that, in expectation, aligns with the complete $n$-gram distribution while markedly reducing variance across mini-batches compared to the standard next-token loss. Empirically, we demonstrate that both the $n$-gram regularized model and our approximation yield substantial improvements in model quality and convergence rate compared to existing methods. Furthermore, our approximation facilitates scalability of gains to larger datasets and models compared to the straightforward $n$-gram regularization method.




\end{abstract}

\section{Introduction}

\begin{figure}[t]
    \centering
    \includegraphics[width=0.5\textwidth]{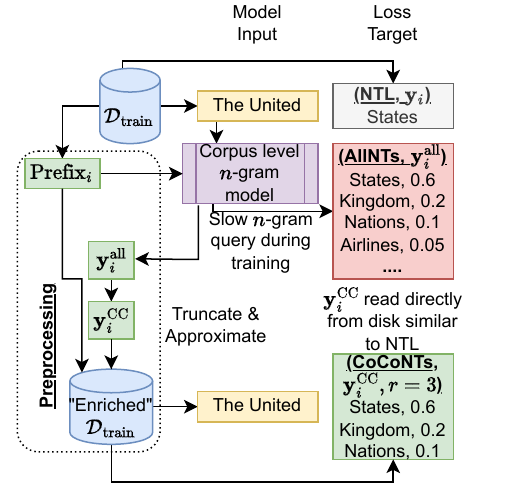}
    \caption{\textbf{Comparison of various training methods.} Standard next-token likelihood reads inputs as well as targets from the disk. $n$-gram augmented methods (AllNTs) obtain targets ($\ymulti{i}$) by querying an $n$-gram model which can be slow during training. Our proposed method, \sysname, truncates and approximates the $\ymulti{i}$ and stores the preprocessed distribution ($\yadj{i}$) along with the dataset itself for faster retrieval during training.}
    \label{fig:concept}
\end{figure}
Since the advent of the first neural language models (NLM)~\citep{bengio2011neural,mikolov2013distributed}, a standard approach to training NLMs has been to maximize the likelihood of the next token (\NT) given the preceding tokens in randomly sampled token sequence from the dataset. 
In this paper we show that LLM pre-training can be made significantly more efficient by supervising  with a distribution over multiple possible next tokens, instead of a single next token.  Earlier studies~\citep{neubig-dyer-2016-generalizing, zhao-etal-2017-ngram2vec, yang-etal-2019-using} have explored using corpus-level $n$-gram statistics to improve the quality of RNN based language models. 
This can also be thought of as label-smoothing \citep{szegedy2016rethinking, muller2019when} with $n$-gram estimated distribution.  


In this paper we show that regularizing with corpus level $n$-gram statistics continues to be beneficial even on current language models, both when pre-training from scratch or fine-tuning.  
However, a major hurdle is scaling up such techniques to today's corpus and model size. 
We address these challenges by 
(1) proposing a compact representation of the next token distribution while continuing to be statistically consistent.
and (2) designing data structures to efficiently make use of these distributions without stalling throughput-optimized linear algebra accelerators (XLA devices) such as TPUs. In particular, we find that our proposed data handling strategy and objective (named ``\sysname'', pronounced ``coconuts'') often matches the performance of $n$-gram models with full (non-truncated) distribution (Sec. \ref{sec:model_performance}). Notably, while the prior $n$-gram augmented methods use storage proportional to the size of training dataset, our storage overheads are \textit{constant} with respect to size of the training dataset (Sec. \ref{sec:training_efficiency}). We also find that $n$-gram augmented training methods (prior works as well as \sysname) can reach the same validation perplexity as the NTL objective in nearly 50\% less optimization steps (Sec. \ref{sec:training_efficiency}). Our proposal is useful for improving model quality in both pre-training (Sec. \ref{sec:babylm}) as well as fine-tuning -- all parameter (Sec. \ref{sec:model_performance}) or parameter efficient (Sec. \ref{sec:peft}).

Our key contributions can be summarized as: (1) We highlight the usefulness of training with n-gram statistics for faster convergence of language models, and propose a statistically  consistent truncation strategy of making their implementation scalable and practical for current data and model sizes~(Sec. \ref{sec:approximation}).
(2) We show how this truncated distribution can be efficiently retrieved during training time without slowing down XLA devices (Sec. \ref{sec:preenriching}, Sec. \ref{sec:minibatch}). (3) We discuss practical strategies and their effects on model performance of scaling this strategy to potentially very large scale datasets via sharding (Sec. \ref{sec:training_efficiency}). (4) We thoroughly test our proposal through ablation studies, comparisons with prior works on both fine-tuning as well as pre-training of language models (Sec. \ref{sec:experiments}).

\newcommand{\vocabsize}{|\vocab|}
\newcommand{\ynt}{\vy^\text{NT}}
\section{Related Work}

\xhdr{Language modeling} 
Early neural language models proposed in \cite{bengio2011neural} and \cite{mikolov2013distributed} were trained to maximize the likelihood of the next token in randomly sampled training batches. While the Transformer architecture \citep{vaswani2017attention} has been the powerhouse of modern LLMs \citep{gpt-neox-20b, zhang2022opt, touvron2023llama}, the key training objective has remained the same. Several studies have explored alternate objectives such as unlikelihood training \citep{Welleck2020Neural} or contrastive loss \citep{su2022a,Jain2023ContraCLM} to improve text generation from these LMs. However,  these objectives are not statistically consistent, and have not been adopted in large scale pre-training. 

\xhdr{Augmenting language model training} 
Some prior work~\citep{mikolov2011extensions, chelba2014billion, jozefowicz2016exploring} have proposed extending RNN based LMs \citep{bengio2011neural, mikolov2013distributed} with KN smoothed $n$-gram LMs, and shown that these result in a better language model. This sparked interest \citep{neubig-dyer-2016-generalizing, zhao-etal-2017-ngram2vec, yang-etal-2019-using} to introduce corpus level $n$-gram statistics into an otherwise local training procedure. \citet{frydenlund2022language} replaced the standard cross-entropy loss with a ranking based loss for which they used $n$-grams as a weak teacher to obtain ``gold'' rankings. \citet{li2020d2gpo} explore additional KL divergence penalty similar to \citet{yang-etal-2019-using} and our work. However, they obtain ground truth distribution by similarity between word embeddings trained on the corpus.

Our work is closest to the $n$-gram regularized loss of \citet{yang-etal-2019-using} but we propose a modified compact supervision that scales to large corpus. 
We also compare our models against \cite{li2020d2gpo} to study whether KL divergence from learned word vectors can serve as a better proxy for relatively expensive count based $n$-gram models.

\section{Language Modeling}
Let $\vocab$ denote a vocabulary of tokens. Given a token sequence $\vt:t_1,t_2\ldots,t_i$ where each $t_j$ in an integer index into the vocabulary $\vocab$, a language model $P_\theta(t | t_1\ldots t_{n}) \in \mathbb{R}^{|\vocab|}$ assigns a multinomial distribution of the probability of the possible next token that could follow $\vt$.  The size of the token sequence $\vt$ is limited to a maximum length $L$.
For training $\theta$ we are given a corpus $\trainset=x_1,\ldots,x_N$ where $N$ is typically very large $N \gg L$.

A standard method of training is to maximize the likelihood of the next token (NTL or NT) given the preceding tokens in a randomly sampled snippet of length $L$ from $\trainset$.   Let $x_{j},\ldots,x_{j+L}$ denote such a snippet sampled at position $j$ of $\trainset$.  The training objective then becomes
\begin{equation}
\label{eq:ntl}
    \hat{\theta}_D = \argmax_\theta \sum_{n=1}^L \log P_\theta(x_{j+n+1} | x_{j},\ldots, x_{j+n})
\end{equation}
When a prefix $\vx_{j:n}=x_{j},\ldots, x_{j+n}$ has multiple occurrences in the corpus, then for the same context depending on the sampling position $j$, the target may be different. But the model $P_\theta$ needs to converge to a consistent distribution.  For example, a prefix ``The United'' could occur multiple times within a corpus with different possible next tokens as shown in Figure~\ref{fig:concept}. If $\vt$ = $t_1,\ldots,t_n$ denotes the tokens in such a prefix, 
it can be seen that at convergence, the next token likelihood  $P_\theta( t | t_{1},\ldots, t_{n})$  for a token sequence  is expected to be equal to the empirical distribution of tokens following $\vt$ over the entire corpus $\trainset$. Let $\ymulti{\vt}$ denote the fractional frequency of the occurrence of tokens of $\vocab$ following all different positions where $\vt$ appears in the corpus $\trainset$.  At convergence we expect:
\begin{equation}
\begin{split}
    &P_{\hat{\theta}_D}( t | \vt) \longrightarrow \ymulti{\vt}, ~~~~\text{where}, \\
&\ymulti{\vt}[w] =\frac{{\sum_{j \in D} \delta(\vx_{j:n}=\vt, x_{j+n+1}=w)}}{\sum_{j \in D} \delta(\vx_{j:n}=\vt)}
\end{split}
\label{eq:conditional}
\end{equation}
In this paper we investigate if the above convergence to the corpus-level next-token distribution can be sped up via changing the training objective to directly match the target distribution $\ymulti{\vt}$.  For long prefixes $\vt$ we do not expect too much repetition, and also maintaining the $\ymulti{\vt}$ proportions for all possible prefixes may incur too much overhead.  So, we fix a maximum prefix length $k$, and instead optimize for a mixture of these two objectives.

\xhdr{The \AllNTL\ objective}
\begin{equation}
\begin{split}
\label{eq:allntl}
      \min_\theta &\sum_{n=k+1}^L   -\log P_\theta(x_{j+n+1} | \vx_{j:n}) \\  
    & + \sum_{n=1}^k \operatorname{KL}{(\ymulti{\vx_{j:n}}; P_\theta(\cdot | \vx_{j:n}))}
\end{split}
\end{equation}
The above loss is reminiscent of the use of the corpus-level $n$-gram statistics to regularize LM training~\citep{yang-etal-2019-using, neubig-dyer-2016-generalizing}.  The second term goes over all $n$-grams $\vt$ of length from 1 to $k$, and attempts to match the learned model distribution to the observed fraction of next tokens in the corpus following that n-gram $\vt$.  

\xhdr{Benefits of \AllNTL} When we supervise the model to match empirical distribution on all possible next tokens after a prefix, the convergence of the model is expected to be faster. In the empirical section we will show that training with even small $n$-grams ($k=4$), gives rise to much lower perplexity for the same computation budget than the original training only for next token likelihood (Eq~\ref{eq:ntl}).  

\xhdr{Overheads of \AllNTL} 
For imposing the \AllNTL\ loss we need to create a data structure like a trie, which for each possible prefix can provide the distribution of next tokens~\citep{heafield-2011-kenlm, heafield-etal-2013-scalable}.  
Querying the trie for every sampled mini-batch is inefficient. These inefficiencies are especially noticeable (Sec. \ref{sec:training_efficiency}) when scaling to larger datasets (>1B tokens) as the batch creation (which includes trie lookup) on CPU is slow.   In the next section we show an alternative method for supervising the next token distribution that significantly reduces these overheads.
\newcommand{\cci}{{\vx_{j:n}}}
\section{Compact and Consistent Next Tokens: \sysname}

We propose to approximate the full empirical next token distribution $\ymulti{\cci}$ with a more compact and consistent supervision $\yadj{\cci}$ at each sampled prefix $\vx_{j:n}$. Unlike $\ymulti{\cci}$ which can be of size as large as the vocabulary size, the alternative we propose is of size at most $r+1$ where $r$ is a chosen hyper-parameter, like 4 or 8 in our experiments.  We design  $\yadj{\cci}$ so that in  expectation over the mini-batches, $\yadj{\cci}$ matches the \AllNTL\ supervision but where the variance across loss terms is significantly smaller than via the supervision in the \NT\ objective.

\label{sec:approximation}
Let $\ytopr_\cci$ denote a truncation of the $\ymulti{\cci}$ where only the top $r$ largest fractions are retained and the rest of the truncated to zero.  
Let $\ohot{\token{i}}$ be one-hot encoding of size $|\vocab|$ with component for $\token{i}$ as 1. We approximate $\ymulti{\cci}$ with $\yadj{\cci}$ as follows:
\begin{equation}\label{eq:approximating}
    \yadj{\cci} = \begin{cases}
        v\ytopr_\cci \quad\quad\quad\quad \text{if }x_{j+n+1} \in \ytopr_\cci\\
        u\ytopr_\cci + \ohot{x_{j+n+1}} \quad\text{otherwise}
    \end{cases}
\end{equation}

\xhdr{Choosing $u,v$}
We choose $u, v$ such that $\mathbb{E}(\yadj{\cci}) \approx {\ymulti{\cci}}$ and $|\yadj{\cci} - \ymulti{\cci}|$ is minimized. Let $p = \sum_t \ytopr[t]$ denote the total probability mass covered by the top-r highest probability tokens.  An example appears in Figure~\ref{fig:concept}. For a token $t \in \vocab$ that is outside the top-r tokens in $\ymulti{\cci}$, it is clear that $\mathbb{E}(\yadj{\cci})[t] = {\ymulti{\cci}}[t]$.  Now consider a token $t$ in the top-r set.  We want to determine values of $u,v$ s.t. $\mathbb{E}(\yadj{\cci})[t] ={\ymulti{\cci}}[t]$ 
\begin{align*}
\mathbb{E}(\yadj{\cci}[t]) & = \ytopr[t](vp + u(1-p)) = \ytopr[t] \\
                     & \implies v = \frac{1 - (1-p)u}{p} 
\end{align*}
This shows that we are "stealing" some probability mass from the top-r token positions and assigning them to the rare token positions, so that across all repetitions of the prefix we have consistent supervision on the next token distribution.  By choosing $u$ carefully we can control this consistency.
The value of $u$ has to be in the range $[0, 1/(1-p)]$ for $\yadj{}$ to be non-zero for all positions.  
The distance $|\yadj{\cci} - \ymulti{\cci}|$ can be computed as 
$$
|\yadj{\cci} - \ymulti{\cci}| \begin{cases}
      = (1-p)u & \quad\text{if }   x_{j+n+1} \in \ytopr_\cci\ \\
    \leq 2-up & \text{otherwise}
\end{cases}
$$
As $u$ increases from zero, the supervision at the rare token positions is made closer to the ideal, but at the cost of the frequent token positions.  We define a hyper-parameter $\gamma > 1$, and choose $u=\frac{1}{\gamma-p}$.  While $p$ varies with prefixes $\vx_{j:n}$ and datasets, we found that a value of $\gamma=1.5$ performs well across diverse settings. We show effectiveness of the approximation via an example in Appendix \ref{sec:appendix_effectiveness}.  


\subsection{Pre-enriching the dataset with $\yadj{}$}
\label{sec:preenriching}
The above design of $\yadj{}$ allows us to tackle the core computational inefficiency of \AllNTL.  Instead of incurring the CPU overheads of trie-lookups during training, we propose to pre-enrich the training corpus with  the compact $\yadj{}$ distributions  stored along with the corpus. 
The total storage cost becomes only $(L+2kr)/L$ times the original storage cost but we avoid the expensive trie lookup operations during training. Also, 
%
we can build the entire trie in-memory only once during the pre-processing phase and discard after the data enrichment phase.  

We implement our trie in C, similar to \citet{heafield-2011-kenlm, heafield-etal-2013-scalable}. Each TrieNode consists of the count and a HashMap where key is the next token ID and value is a pointer to the next TrieNode. The HashMap is implemented using an AVL tree. We implement a Top-$r$ query method which returns Top-$r$ token IDs and Top-$r$ probabilities for a given prefix of upto $k$ tokens using Eq. \ref{eq:conditional}.
 Concretely, we first construct the trie by reading sequences of $k$ tokens from the dataset and inserting it in the trie. At every level $i$, we increment the count by 1 to implicitly record prefix $\prefix{i}$. Once the trie is constructed, we start reading (disjoint) sequences of $L$ tokens from the dataset and writing the ``enriched'' sequence back to disk. To ``enrich'' a sequence $x_j,\ldots,x_{j+L}$, we first look up the prefix in the trie and get pointers to $k$ nodes i.e. one at each level. At each of these nodes, the HashMap is storing $\ymulti{}$. We can efficiently traverse this HashMap in descending order of fractions to get $\ytopr$. Once we get $\topr{r}{\ymulti{j:i}}\; \forall i \in [0,k)$, we can write the ``enriched'' sequence to disk. Note that, this operation still requires disk storage of $(L + (L + 2kr))\times N$ tokens where $N$ is the total number of sequences. We present additional discussion and a sample walkthrough of this procedure for more clarity in Appendix \ref{sec:appendix_preenriching}. 

\subsection{Building the mini-batch with $\yadj{i}$}
\label{sec:minibatch}
To build the mini-batch of size $B$ in standard NTL, one simply reads sequences of $L$ tokens $B$ times and concatenates them. In standard NTL, the sequence of $L$ tokens itself is both input and target but in \sysname, we have 2 targets that need to be built. In \sysname\, we need to read sequence of $L + 2kr$ tokens from the disk and use the first $L$ tokens to form $\vx, \vy$ similar to NTL. Next $2r$ ``tokens'' correspond to $\ytopr$ where the first token is the actual token ID and the next token is really the count of that token appearing after prefix $[\token{1}]$. The $2r$ tokens following this would correspond to $\topr{r}{\ymulti{2}}$ which encodes counts and tokens appearing after prefix $[\token{1}, \token{2}]$. This would repeat for $k$ times to give $k$ distributions as targets for KL divergence in Eq. \ref{eq:allntl}. We provide additional discussion on memory footprint in Appendix \ref{sec:appendix_minibatch}.

\section{Experiments}

Through the experiments, we empirically validate the effectiveness of our approximation (Sec. \ref{sec:model_performance}) along with the ablation studies on various hyperparameters. We also evaluate the training efficiency of \sysname\ in Sec. \ref{sec:training_efficiency}. Finally, we present two case studies on \sysname\ in pre-training (Sec. \ref{sec:babylm}) and parameter-efficient fine-tuning (Sec. \ref{sec:peft}) to further demonstrate the usefulness and relevance of \sysname\ even in modern LLM usecases. Hyperparameters for all the experiments are presented in Appendix \ref{sec:appendix_hparams}.

\label{sec:experiments}

\subsection{Datasets, Baselines and Metrics}
We explore the effectiveness of \sysname\ in fine-tuning existing models on WikiText-103 \citep{merity2017pointer}, MiniPile \citep{kaddour2023minipile, gao2020pile} and a subset of PubMed-Abstracts \citep{luo2022biogpt}. The WikiText-103 training split consists of $\approx 114$M tokens while MiniPile and PubMed splits consist of $\approx$ 1.6B and 2.6B tokens respectively.

For our full fine-tuning experiments, we compare \sysname\ against \AllNTL\ \citep{yang-etal-2019-using} and NTL objectives on WikiText-103. We compare against D2GPO baseline \citep{li2020d2gpo} as well since they also use a KL divergence based augmentation of training loss. 

Since \citet{su2023contrastive} raised concerns about isotropy of the base gpt2-125m \citep{radford2019language} model, we also study effects of \sysname\ objective on gpt-neo-125m \citep{black2021gptneo} and opt-125m, opt-1.3B \citep{zhang2022opt}.

\xhdr{Metrics} Following \cite{su2022a}, we evaluate each fine-tuning method several on model quality and text quality metrics. 
\begin{itemize}[leftmargin=0.4cm, itemsep=0pt]
    \item Perplexity (\textbf{ppl}) of the model on the test set.
    \item Prediction accuracy (\textbf{acc}) of the model. Given a sample with inputs $\seqx{L}$ and labels $\seqy{L}$ from the test set, we take argmax of each of the $L$ predicted distribution at time step $t$ to get top-1 predicted token and match it against $\y{t}$ to calculate the prediction accuracy. 
    \item Repetition (\textbf{rep}) measures the fraction of top-1 next-token predictions that occur in the prefix.
    \item Expected calibration error (\textbf{ECE}) measures how over/underconfident is the model when making correct/incorrect predictions. Lower ECE indicates better calibrated models.
    \item \textbf{MAUVE (MVE)}~\citep{pillutla2021mauve} measures
    the similarity between the generated text and reference text using the embeddings of another large pretrained model.
    \item Repetition within a generated single text sequence: (\textbf{rep-n})  $100\times(1 - \frac{|\text{unique } n\text{-grams}|}{|\text{total } n\text{-grams}|})$.
    \item Diversity (\textbf{div.}) measures repetition at different $n$-gram levels: $\prod_{n=2}^4(1 - \frac{\text{rep-}n}{100})$.
    \item Number of unique bigrams (\textbf{\#uniq}) in the generated text.
    
\end{itemize}

We also compare the Zipf coefficient of the generated text to gold text as suggested by \citep{Meister2023}. For generating text, we either use greedy decoding or nucleus \citep{holtzman2019curious} sampling. To compare training efficiency, we use following metrics:

\begin{itemize}[leftmargin=0.4cm, itemsep=0pt]
    \item Number of optimization steps (\textbf{steps-to-ppl}) taken to reach NTL's perplexity on the val set.
    \item Total wall clock time (\textbf{TWT}) to finish the entire training. We exclude the time for preprocessing (trie building and storage) as the preprocessing is a one-time operation which many times did not take very long. Additionally, one could always start with datasets that are preprocessed by someone else. We do include the time it takes to load the trie and training time retrieval as these operations will often stall the XLA devices. 
    \item Total disk usage (\textbf{disk}) required for training (includes storage of prefix trie).
    \item Maximum CPU RAM (\textbf{max-RAM}) used for pretraining i.e. loading and using the trie.
\end{itemize}

\begin{table*}[!ht]
    \centering
    \footnotesize
    \begin{tabular}{lrrrrrrrrrrr}
        \toprule
        \multirow{2}{*}{Model} & \multicolumn{4}{c}{Model Quality} & \multicolumn{3}{c}{Generations [greedy]} & \multicolumn{4}{c}{Generations [nucleus]} \\
        \cmidrule(lr){2-5} \cmidrule(lr){6-8} \cmidrule(lr){9-12}
         & PPL & Acc. & Rep. & ECE & Div. & \#Uniq & Zipf & Div. & \#Uniq & Zipf & MVE \\ 
        \midrule
\multicolumn{12}{c}{gpt2-125m}\\
NTL & 24.279 & 39.667 & 52.597 & 0.075 & 14.051 & 87756 & 0.990 & 94.504 & 112768 & 0.952 & 0.708 \\ 
D2GPO & 22.151 & \cellcolor{yellow!25}42.540 & 52.270 & \cellcolor{green!15}0.068 & 14.823 & 93563 & \cellcolor{green!15}0.945 & 95.289 & 122590 & 0.866 & 0.774 \\ 
\AllNTL  & \cellcolor{green!15}20.634 & \cellcolor{green!15}44.170 & \cellcolor{green!15}48.922 & \cellcolor{green!15}0.068 & \cellcolor{green!15}30.678 & \cellcolor{green!15}127266 & \cellcolor{yellow!25}0.958 & \cellcolor{green!15}99.415 & \cellcolor{green!15}161138 & \cellcolor{green!15}0.911 & \cellcolor{green!15}0.900 \\ 
\sysname\  & \cellcolor{yellow!25} 20.717 & 42.192 & \cellcolor{yellow!25}50.300 & \cellcolor{yellow!25}0.069 & \cellcolor{yellow!25}30.030 & \cellcolor{yellow!25}121498 & 0.959 & \cellcolor{yellow!25}95.751 & \cellcolor{yellow!25}156432 & \cellcolor{yellow!25}0.932 & \cellcolor{yellow!25}0.863 \\ 
\midrule
\multicolumn{12}{c}{gpt-neo-125m}\\
NTL & 22.746 & 43.528 & 49.843 & 0.055 & 28.937 & \cellcolor{yellow!25}99326 & 0.956 & \cellcolor{yellow!25}96.809 & 118061 & 0.976 & 0.636 \\ 
\AllNTL  & \cellcolor{green!15}20.188 & \cellcolor{green!15}46.308 & \cellcolor{green!15}47.632 & \cellcolor{green!15}0.049 & \cellcolor{green!15}32.603 & \cellcolor{green!15}101725 & \cellcolor{green!15}0.927 & 96.066 & \cellcolor{green!15}121729 & \cellcolor{green!15}0.910 & \cellcolor{yellow!25}0.685 \\ 
\sysname\  & \cellcolor{yellow!25}20.208 & \cellcolor{yellow!25}44.147 & \cellcolor{yellow!25}49.014 & \cellcolor{yellow!25}0.051 & \cellcolor{yellow!25}31.178 & 99080 & \cellcolor{yellow!25}0.950 & \cellcolor{green!15}96.919 & \cellcolor{yellow!25}120084 & \cellcolor{yellow!25}0.942 & \cellcolor{green!15}0.699 \\ 
\midrule
\multicolumn{12}{c}{opt-125m}\\
NTL & 19.374 & 44.583 & 51.707 & 0.067 & 26.789 & 90484 & 1.022 & 94.771 & 114472 & 0.978 & 0.648 \\ 
\AllNTL  & \cellcolor{yellow!25}19.371 & \cellcolor{green!15}49.282 & \cellcolor{green!15}50.027 & \cellcolor{yellow!25}0.054 & \cellcolor{yellow!25}33.767 & \cellcolor{yellow!25}95887 & \cellcolor{yellow!25}0.994 & \cellcolor{green!15}97.690 & \cellcolor{yellow!25}135104 & \cellcolor{yellow!25}0.951 & \cellcolor{yellow!25}0.670 \\ 
\sysname\  &  \cellcolor{green!15}18.558 & \cellcolor{yellow!25}49.059 & \cellcolor{yellow!25}50.071 & \cellcolor{green!15}0.052 & \cellcolor{green!15}34.722 & \cellcolor{green!15}99168 & \cellcolor{green!15}0.972 & \cellcolor{yellow!25}95.111 & \cellcolor{green!15}141256 & \cellcolor{green!15}0.934 & \cellcolor{green!15}0.703 \\ 
\midrule
\multicolumn{12}{c}{opt-1.3B}\\
NTL & 16.554 & 49.686 & \cellcolor{green!15}45.617 & 0.052 & 29.137 & 101649 & \cellcolor{green!15}0.939 & 97.723 & 164789 & \cellcolor{green!15}0.912 & 0.861 \\ 
\sysname\  & \cellcolor{green!15}15.679 & \cellcolor{green!15}50.010 & 45.617 & \cellcolor{green!15}0.051 & \cellcolor{green!15}29.866 & \cellcolor{green!15}102492 & 0.941 & \cellcolor{green!15}98.111 & \cellcolor{green!15}166846 & 0.941 & \cellcolor{green!15}0.869 \\ 
\midrule
Gold & ~ & ~ & ~ & ~ & 89.036 & 171076 & 0.925 & 89.036 & 171076 & 0.925 & 1.000 \\ 
\bottomrule\\
    \end{tabular}
    \caption{\textbf{Results on WikiText-103.} We find that \sysname\ is competitive with far more expensive (as we show in Sec. \ref{sec:training_efficiency}) \AllNTL\ objective. For both \AllNTL\ and \sysname, $k=8$ was used to build the prefix trie. \sysname\ additionally used $r=8$. Best results are highlighted with green while second-best are highlighted with yellow.}
    \label{tab:wikitext103}
\end{table*}
\subsection{Model performance}\label{sec:model_performance}
\xhdr{\sysname\ is comparable to \AllNTL\ and better than NTL} In Table \ref{tab:wikitext103}, we compare performance of \sysname\ with various other objectives. We find that \sysname\ is able to provide consistent gains over the NTL baseline. We also notice that \sysname\ outperforms D2GPO \cite{li2020d2gpo} which indicates that count based conditional $n$-gram models are able to provide a stronger training signal as compared to word embedding similarity. Maximum gains with \sysname\ are observed on gpt2-125m, however this could be related to isotropy of gpt2-125m checkpoint as discussed by \citep{su2023contrastive}. Across all metrics and models, \AllNTL\ is generally the best performing model while \sysname\ comes close to it. \sysname\ offers gains (albeit modest as compared to small models) to larger models (opt-1.3B) as well. 

\subsection{Training efficiency}
\label{sec:training_efficiency}

\begin{figure*}[t]
    \includegraphics[width=\textwidth]{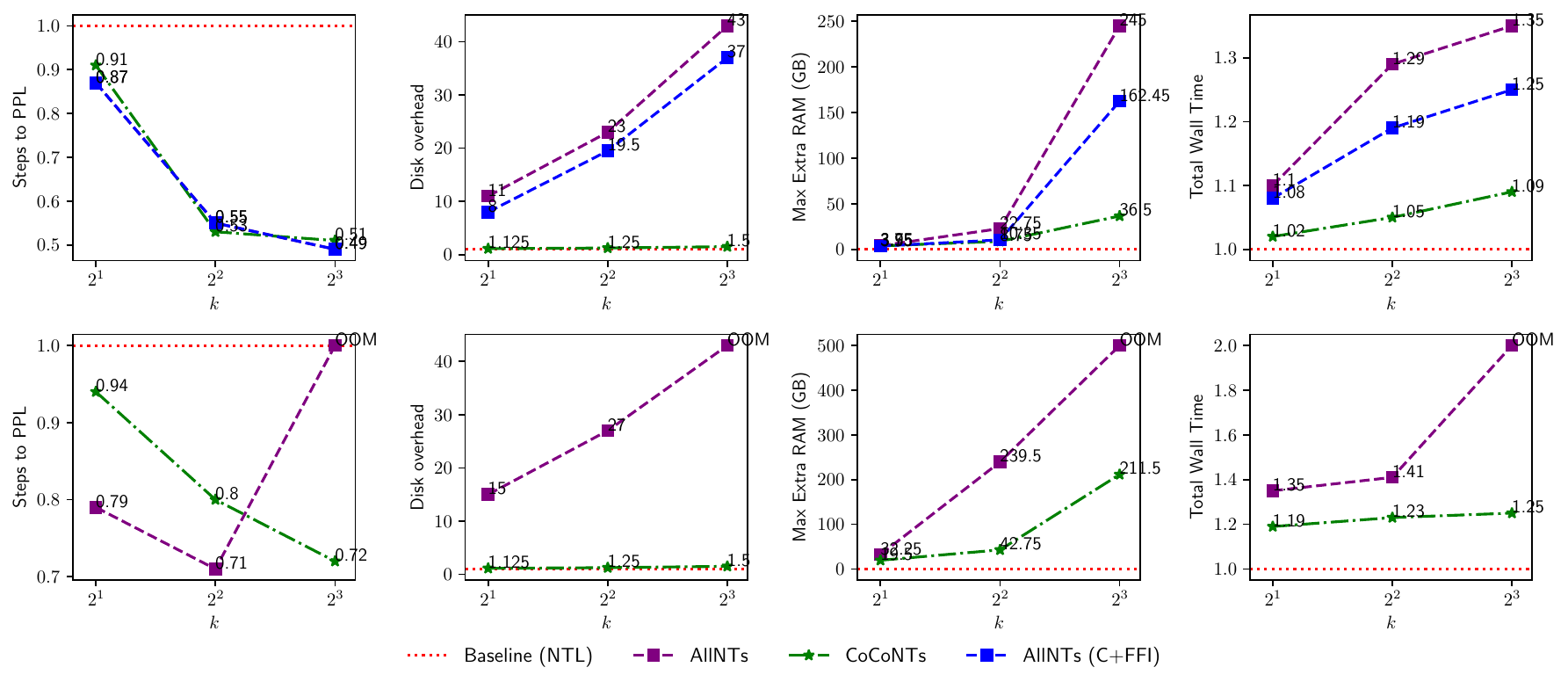}
    \caption{\textbf{Comparison of training efficiency on WikiText-103 (top) and PubMed (bottom).} \AllNTL\ with higher values of $k$ can easily go out of memory from a naive implementation. Both \AllNTL\ and \sysname\ converge faster to NTL's validation perplexity as compared to NTL. The total wall time (TWT) to finish the entire training is also significantly lower with \sysname\ as compared to \AllNTL\ due to lack of any $n$-gram querying during training. gpt2-125m model is used for all experiments with $r=8$ for \sysname.}
    \label{fig:efficiency}
\end{figure*}

\xhdr{\sysname\ is significantly more efficient than \AllNTL} Figure \ref{fig:efficiency} compares our training efficiency metrics across various training methods on WikiText-103 and PubMed datasets. \AllNTL\ uses Python's \texttt{defaultdict} \footnote{\tiny{\url{https://docs.python.org/3/library/collections.html\#collections.defaultdict}}} to implement the trie HashMap while \sysname\ uses our efficient implementation of HashMap as described in Sec. \ref{sec:preenriching}. \AllNTL\ serializes the resultant trie to disk using the Python's \texttt{pickle}\footnote{\tiny{\url{https://docs.python.org/3/library/pickle.html}}} library. We also explore using an existing $n$-gram implementation \cite{heafield-2011-kenlm} with Python FFI as denoted by \AllNTL-CFFI. Because \AllNTL\ stores the $n$-gram model on disk for later retrieval, its disk overhead grows with the size of the dataset as opposed \sysname's constant (dependant on $k,r$ only) disk overhead. Measuring the total wall clock time, we find that \sysname\ ($k=4$) is \textit{faster} than \AllNTL\ ($k=2$) which highlights the benefits of our approximation and pre-enriching of the dataset. 

Results on PubMed demonstrate challenges in scaling up vanilla \AllNTL\ to large datasets. On our machine with 256GB RAM, the prefix trie for $k=8$ with vanilla \AllNTL\ did not fit in memory. It is possible that \AllNTL-CFFI is able to fit everything in memory similar to \sysname\ but we could not explore in depth since our $n$-gram implementation \cite{heafield-2011-kenlm} kept crashing on our system. While \AllNTL-CFFI saves disk and RAM, it still is not as efficient as \sysname\ on WikiText-103. Moreover, the trie loading time as well as the (somewhat) slow CFFI interface serialization overheads significantly increase total wall time of \AllNTL-CFFI over \sysname. 


\begin{figure*}
  \begin{minipage}[c]{0.65\textwidth}
    \includegraphics[width=\textwidth]{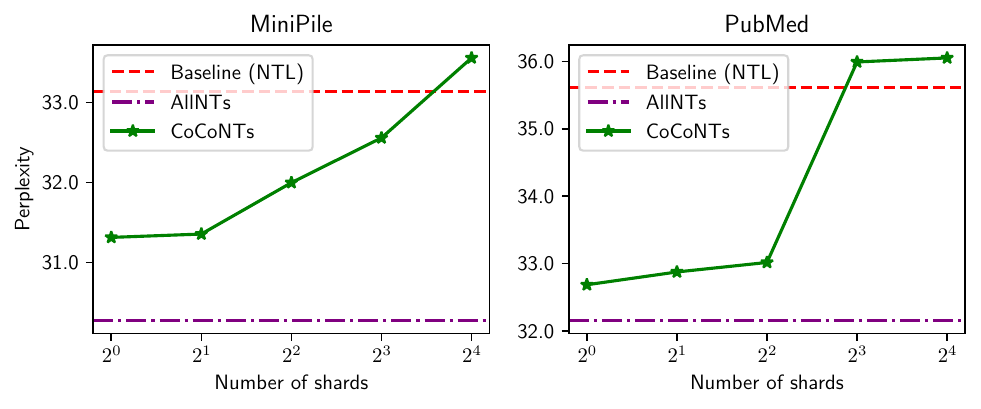}
  \end{minipage}\hfill
  \begin{minipage}[c]{0.3\textwidth}
    \caption{
       \textbf{Effect of sharded \sysname\ on large datasets.} Oversharding can make the $n$-gram distribution unreasonably sparse. This can lead to overly optimistic approximation and KL penalty which can hurt the performance on extremely small indices.
    } \label{fig:sharding}
  \end{minipage}
\end{figure*}

\xhdr{Effects of sharding} As compared to datasets used in this work, modern LLMs \citep{gpt-neox-20b, touvron2023llama, groeneveld2024olmo} are often trained on far bigger web corpora \citep{gao2020pile, together2023redpajama,dolma} for which building an $n$-gram model in-memory may not be feasible. In such cases, we show that such datasets can be sharded into multiple small datasets of several few billion tokens with each shard being enriched with its own $n$-gram index. We study the effect of sharding by simulating it on MiniPile and PubMed datasets and seeing effect on perplexity as shown in Fig. \ref{fig:sharding}. We find that after a certain threshold of shards, the number of tokens per shard decreases so much that KL penalty can become overly optimistic and result in worse perplexity. In general, we found that having more than billion tokens per shard was sufficient to get results close to \AllNTL\ while still using modest amount of RAM.

\subsection{Understanding \sysname}
\label{sec:ablations}
\begin{figure*}[t]
  \begin{minipage}[c]{0.65\textwidth}
    \includegraphics[width=\textwidth]{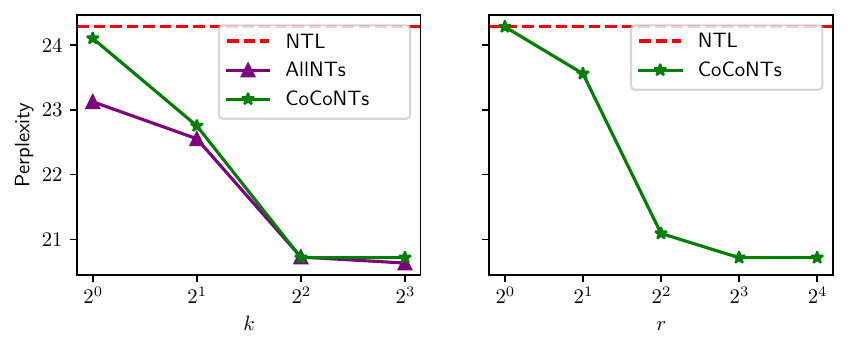}
  \end{minipage}\hfill
  \begin{minipage}[c]{0.3\textwidth}
    \caption{
       \textbf{Ablations studies on $k$ and $r$ for \AllNTL\ and \sysname.} All experiments fine-tune a gpt2-125m model on WikiText-103. Higher values of both $k$ and $r$ improve perplexity before plateauing. $r=8$ is fixed when varying $k$ and $k=8$ is fixed when varying $r$ for \sysname.
    } \label{fig:ablations}
  \end{minipage}
\end{figure*}
\xhdr{Ablation studies on $k$ and $r$} By default we choose $k=8,r=8$. We report ablations on these values  when fine-tuning gpt2-125m on WikiText-103 and compare using validation perplexity. As indicated by trends in Fig. \ref{fig:ablations}, we find that increasing $k$ and $r$ leads to predictable improvements in perplexity as compared to the NTL baseline. We do notice a significant difference in perplexities between \AllNTL\ and \sysname\ for lower values of $k$. This could be potentially due to poorer $\yadj{i}$ approximation since the fan-out is expected to be significantly higher for initial few tokens. Empirically, in the first few levels of the trie, the branching factor is the highest often leading to $\ymulti{i}$ that has a support size much larger than $r$. This is further supported by the significant improvement observed when going from $r=2$ to $r=4$ for a fixed $k$. Moreover, as $k$ increases, the support of $\ymulti{i}$ naturally decreases. In fact for $k > 2$, the average support size is less than 4 on WikiText-103. This implies that $p=1$ in Eq. \ref{eq:approximating} leading to $v=1$ which effectively reduces \sysname\ to \AllNTL\ objective.


\begin{table}[t]
\centering
\begin{tabular}{lrrr}
\toprule
& Wiki-103 & MiniPile & PubMed \\
\midrule
gpt2-125m               & 24.279       & 43.456   & 37.819 \\
\AllNTL\                  & 20.634       & 41.261   & 38.538 \\
\sysname\             & 20.717       & 42.486   & 39.016 \\
$\quad$+MiniPile   & \textbf{20.223}       & \textbf{36.899}   & 38.564 \\
$\quad$+PubMed     & 20.926       & 39.432   & \textbf{36.175}  \\
\bottomrule
\end{tabular}
\caption{\textbf{Perplexity improvements when augmenting $n$-gram index with larger/domain specific data.} Each model (row) is trained on WikiText-103 and evaluated on validation splits of (column) WikiText-103, MiniPile and PubMed. \sysname\ with indices built on additional MiniPile or PubMed data improves performance on respective datasets.}
\label{tab:extra-dataset-ablation}
\end{table}

\xhdr{Using larger/domain specific data sources to build $n$-gram models can help} We show that using  $n$-gram statistics from alternative corpus is also useful. To study this, we augmented our existing WikiText-103 $n$-gram trie with $n$-grams from MiniPile and PubMed datasets independently. As shown in Table \ref{tab:extra-dataset-ablation}, model trained with  WikiText-103 + MiniPile index improves perplexity on both WikiText-103 \textit{and} MiniPile. On the other hand, if the augmenting dataset (PubMed) is both large \textit{and} domain specific, the resultant model improves 
on augmenting 
dataset \textit{at the cost} of performance on the original (WikiText-103) dataset.


\begin{figure}[t]
    \centering
    \includegraphics[width=0.5\textwidth]{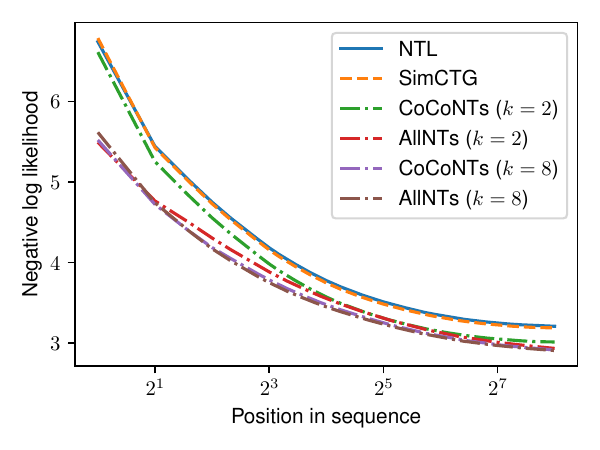}
    \caption{\textbf{Perplexity as a function of position in sequence.} Both AllNTs and CoCoNTs show smooth changes in perplexity despite applying loss to only a small $k$ token prefix.}
    \label{fig:nll_vs_seq_len}
\end{figure}

\xhdr{Perplexity reduction as a function of prefix length}
We show that even though we impose the \sysname\ loss only in a small prefix of an overall sequence of length 256, the improvement in model quality is throughout the length of the sequence. For this, we measure the negative log-likelihood (NLL) separately at each position from 1 to 256 in the test data and show our findings in Figure~\ref{fig:nll_vs_seq_len}.
We observe that we reduce NLL at all sequence positions over the two existing methods.  
This suggests that in today's NLM architectures with shared parameters, enforcing the correct loss in only a subset of the positions is sufficient for overall gains.

\begin{table}[t]
    \centering
    \small
    \begin{tabular}{lrrrrrr}
        \toprule
        \multirow{2}{*}{Method} & \multicolumn{6}{c}{Pretraining Step}\\
        \cmidrule(lr){2-7}
         & 10k & 20k & 30k & 40k & 50k & 60k \\ 
        \midrule
        NTL & 26.3 & 23.9 & 23.0 & 22.7 & 22.5 & 22.4 \\ 
        AllNTs & 25.5 & 22.4 & 21.2 & 20.2 & 20.2 & 20.0 \\ 
        CoCoNTs & 25.9 & 22.6 & 21.3 & 20.2 & 20.2 & 20.1 \\ 
        \bottomrule
    \end{tabular}
    \caption{\textbf{Effect of pretraining for more steps on validation perplexity.} Both AllNTs and CoCoNTs use $k=8$. CoCoNTs also uses $r=8$. The difference in CoCoNTs and the baseline slightly reduces but does not become completely zero.}
    \label{tab:pretraining_longer}
\end{table}

\xhdr{Effect of longer pretraining} We study that pretraining for more steps flattens the benefits of CoCoNTs slightly but not completely. We continued pretraining of gpt-neo-125m from our experiments (Sec. \ref{sec:experiments}) for 20k more steps (50\% more than original pretraining steps) and report our findings in Table \ref{tab:pretraining_longer}. Looking at the validation perplexities, we do see that the difference between baseline and CoCoNTs perplexities does decrease from 2.5 to 2.3 over the 20k additional steps. To correctly assess if this difference goes to zero, significantly longer training is required.

\subsection{Case Study: PEFT Domain Adaptation} 
\label{sec:peft}
Parameter efficient fine-tuning (PEFT) of LLMs \cite{hu2022lora, dettmers2023qlora} has become a popular method to adapt general purpose LLMs such as LLaMA-2 \cite{touvron2023llama} on a specific domain. To study benefits of \sysname\ objective in this direction, we first train a LoRA \cite{hu2022lora} to fine-tune LLaMA-2-7B on our split of the PubMed dataset using each of the fine-tuning method. Both \AllNTL\ and \sysname\ use $k=4$ while \sysname\ uses $r=8$. With this LoRA as the starting point, we fine-tune the resultant (Medical LLaMA) model on 2 medical QA (PubMedQA \cite{jin-etal-2019-pubmedqa} and MedQA \cite{jin2021disease}) tasks independently. As summarized by accuracy of these QA tasks in Table \ref{tab:llama-lora-results}, we find results similar to full fine-tuning i.e. \sysname\ performs better than NTL but worse than \AllNTL. However, \AllNTL\ incurs an almost 60\% increase in pre-training time, whereas our methods reduces the overheads by half as seen by the last total wallclock time (TWT) column in the table.  

\begin{table}[t]
\centering
\begin{tabular}{lrrr}
\toprule
            & PubMedQA & MedQA  & TWT\\
\midrule
NTL  & 56.025   & 36.911 & 1x \\
\midrule
\AllNTL\      & \textbf{56.976}   & \textbf{38.343} & 1.57x\\
\sysname\ & 56.251   & 37.732 & \textbf{1.32x} \\
\bottomrule
\end{tabular}
\caption{\textbf{Results with LoRA domain adaptation.} The base LLM (LLaMA-2-7B) is finetuned on PubMed abstracts and subsequently finetuned on each of the QA tasks with LoRA. \sysname\ continues to outperform NTL while staying close to \AllNTL. The Total Wallclock Time (TWT) of \AllNTL\ is much higher than ours.}
\label{tab:llama-lora-results}
\end{table}

\begin{table}[t]
\centering
\begin{tabular}{lrrrr}
\toprule
            & CoLA  & MRPC  & RTE & TWT   \\
            & (MCC)  & (F1)  & (Acc)   \\
\midrule
NTL    & 0.288 & 0.735 & 0.591 & 1x\\
\midrule
\AllNTL      & \textbf{0.347} & \textbf{0.771} & \textbf{0.629} & 1.35x\\
\sysname\ & 0.339 & 0.742 & 0.621 & \textbf{1.19x}\\
\bottomrule
\end{tabular}
\caption{\textbf{Results on the BabyLM Strict Challenge.} The base model (opt-125m) trained with \sysname\ performs similar to \AllNTL\ while being significantly faster than \AllNTL, as measured by the Total Wallclock Time (TWT). Both methods are better than NTL on quality.}
\label{tab:babylm-results}
\end{table}

\subsection{Case Study: The BabyLM Challenge}
\label{sec:babylm}
Can \sysname\ be used to pre-train a better language model for downstream tasks? We pre-train an opt-125m architecture model from scratch on the data from the BabyLM challenge \cite{warstadt-etal-2023-findings}. We use the nearly 100M word data from the ``strict'' track with standard preprocessing. Once all (NTL, \AllNTL, \sysname) the models are pre-trained, we finetune each on 3 downstream tasks (with recommended hyperparameters) from the evaluation suite -- CoLA \cite{warstadt-etal-2019-neural}, MRPC \cite{dolan-brockett-2005-automatically} and RTE \cite{dzikovska-etal-2013-semeval} which are subsets of the SuperGLUE benchmark \cite{wang2018glue, wang2019superglue} for the BabyLM challenge. Task specific fine-tuning does not use any custom objective. Table \ref{tab:babylm-results} show metrics on respective tasks. 
Both \AllNTL\ and \sysname\ trained base LMs show significant improvement over standard NTL in CoLA which is about linguistic acceptability. This could be due to grammatically incorrect sentences/completions automatically being suppressed in the $n$-gram index. On other downstream tasks such as paraphrase detection (MRPC) or entailment (RTE), the performance of all models is much closer in comparison.  Also, observe that the total wall clock time (TWT) for \AllNTL\ is 35\% higher than baseline, and \sysname\ reduces the overhead down to 19\%.

\section{Conclusion}


In this work, we revisited the benefits of regularizing language model training with corpus-level $n$-gram statistics, and  proposed ways to scale up their implementation on current scales of data and model sizes.  
Our proposal truncates the $n$-gram estimated next-token distribution and introduces a novel method of mixing with occurrences of frequent and rare tokens so as to provide low-variance supervision of the distribution of next tokens. The distributions are designed to be compact and can be stored with the corpus so that their retrieval is as simple as a disk read operation. 
We empirically show that \sysname\ performs comparable to \AllNTL\ objective  but is significantly more efficient 
than \AllNTL. Notably, while the \AllNTL\ storage costs scale with dataset, \sysname\ storage costs depend only on the hyperparameters $k$ and $r$. Our fine-tuning experiments suggest that \sysname\ based training benefits smaller models the most with larger models seeing only modest improvements. We also observe that imposing \AllNTL\ or \sysname\ loss only on a small $k$ token prefix was sufficient to improve the overall model performance. Case study on the BabyLM challenge highlights that \sysname\ trained base LMs are better than standard NTL trained base LMs on downstream tasks as well.




\section*{Limitations and Future Work}




 In this section, we discuss the limitations of \sysname\ objective and provide insights into potential challenges and areas for improvement. The one-time preprocessing step required by our method  for very large-scale datasets requires sharding. Effects of such sharding on social biases of the model must be studied carefully. 
%
When applying our method to very large datasets like C4 or The Pile, the implementation of the prefix-trie using better optimized libraries as discussed in \cite{Jurasfsky2023} may become necessary. This could require significant engineering efforts to optimize access times and ensure efficient training. To address scalability concerns further, a possible suggestion in addition to sharding would be to ``sparsify'' the trie for such large-scale datasets. By pruning branches with low counts, we can significantly reduce the overall memory footprint while still maintaining the essence of (idea of ``heavy hitters'' \citep{misra1982heavy,woodruff2016new,braverman2017bptree}) the next token distribution. 

When applying our method to very large datasets like C4 or The Pile, the implementation of the prefix-trie using better optimized libraries as discussed in \cite{Jurasfsky2023} may become necessary. This could require significant engineering efforts to optimize access times and ensure efficient training. To address scalability concerns further, a possible suggestion in addition to sharding would be to ``sparsify'' the trie for such large-scale datasets. By pruning branches with low counts, we can significantly reduce the overall memory footprint while still maintaining the essence of  the empirical next token. 

Furthermore, it is important to acknowledge that our \sysname\ objective aims to match the empirical next-token distribution, and thus inherits any biases present in the training data. However, an advantage of our approach is that the prefix trie allows for detailed exploration and identification of these biases. If such biases are observed, it should be possible to edit the prefix trie to mitigate their influence. 
Our experimental setup included the largest models and datasets that could run comfortably on our compute resources. Future work can explore effectiveness of \sysname\ on even larger datasets (e.g. ThePile \cite{gao2020pile}, 
RedPajama \cite{together2023redpajama}, Dolma \cite{dolma} etc.) and larger scale models such as LLaMA2 \cite{touvron2023llama}.

\section*{Ethics Statement}

We acknowledge that our objective entails preprocessing and handling large-scale datasets for creating the prefix trie. This necessitates careful attention to privacy concerns and the implementation of robust data protection measures. It is vital to thoroughly examine and mitigate any biases that may emerge in the training data prior to the application of our proposed objective.

Moreover, given the improved text generation capabilities demonstrated by our approach, it is imperative to address ethical considerations regarding the responsible use of language models trained using our proposed objective. In this context, we underscore the significance of ensuring that the deployment and utilization of such models align with ethical standards, including but not limited to mitigating the potential for malicious use, promoting fairness in algorithmic decision-making, and safeguarding user privacy.

\bibliography{anthology,custom,refs}
\bibliographystyle{acl_natbib}

\clearpage

\appendix

\section{Appendix}
\label{sec:appendix}

\subsection{Hyperparameters}
\label{sec:appendix_hparams}
Most of our experiments are performed on a single TPUv2-8 core VM with TPU VM architecture. We also sometimes used 4x NVIDIA A100 GPUs with FlashAttention for a fraction of all experiments. 

For fine-tuning experiments (Sec. \ref{sec:model_performance}), we start with publicly available checkpoints for gpt2-125m\footnote{\tiny{\url{https://huggingface.co/gpt2}}}, gpt-neo-125m\footnote{\tiny{\url{https://huggingface.co/EleutherAI/gpt-neo-125m}}}, opt-125m\footnote{\tiny{\url{https://huggingface.co/facebook/opt-125m}}} and opt-1.3B\footnote{\tiny\url{https://huggingface.co/facebook/opt-1.3B}}. Each 125m parameter model is trained for 40k steps with AdamW \cite{loshchilov2018decoupled} optimizer with effective batch size of 192. The maximum learning rate was set to $10^{-4}$ with 10\% of maximum steps as warmup followed by cosine decay to zero. For the 1.3B parameter model, we set the maximum steps to 10k and reduce the batch size to 32. Each fine-tuning run took roughly 5-6 hours on TPUs and 8-10 hours on GPUs.

For pre-training on the BabyLM challenge (Sec. \ref{sec:babylm}), we set the batch size, optimizer and learning rate schedule similar to fine-tuning and trained for total of 5 epochs. For downstream tasks, we use the hyperparameters mentioned in the BabyLM evaluation pipeline\footnote{\tiny{\url{https://github.com/babylm/evaluation-pipeline}}}. The pre-training took 5.5 hrs on TPUs while fine-tuning on downstream tasks took 1-1.5hrs each.

For parameter-efficient fine-tuning of LLaMA (Sec. \ref{sec:peft}), we set the batch size to 32 and fine-tune on PubMed with hyperparameters similar to WikiText-103 fine-tuning. We use the same parameters as BabyLM downstream tasks for LLaMA PEFT downstream tasks as well. The pre-training took close to 6 hours on TPUs while fine-tuning on downstream tasks took 2-2.5hrs each.

\subsection{Effectiveness of $\yadj{}$}
\label{sec:appendix_effectiveness}
\xhdr{Example} We show a simple example to illustrate how $\yadj{}$ manages to reduce variance in the next-token distribution across sampled mini-batches. Assume vocab size is 5, and $\ymulti{\vt}=[0.6,0.3,0.1,0.05,0.05]$. Let $r=2$.  Assume $\gamma=1.5$.  For this case we have $p=0.9, u=1.66,v=0.5$. Everytime we sample a mini-batch where next token is from the top-2 set: $\{1,2\}$, we supervise with $\yadj{}=v[0.6,0.3,0,0,0]$ which has a distance of 0.166 from $\ymulti{}$. Whereas, when we sample the last three tokens we are at most 0.51 from $\ymulti{}$.  Contrast this to the baseline NT case where for rare token the distance to $\ymulti{}$ could be as high as 1-0.05=0.95!  With $\yadj{}$ we reduce this distance to 0.51.  Even for frequent tokens the distance has been reduced from a maximum of 0.7 to 0.166. 

\begin{figure}[t]
    \centering
    \includegraphics[width=0.5\textwidth]{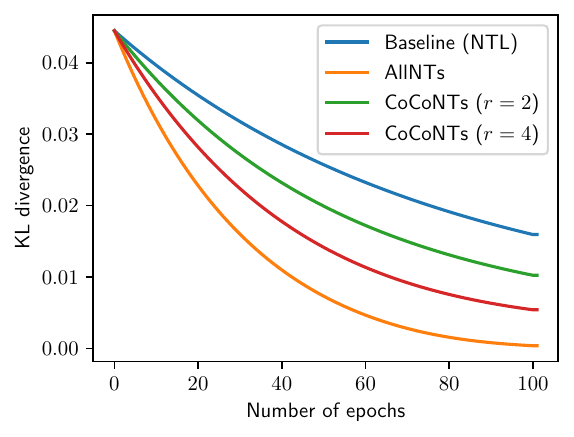}
    \caption{\textbf{Studying convergence rates of our approximation when learning a single 10-class multinomial.} The trajectories are averaged over 10 independent runs for all methods. Higher values of $r$ yield better approximations as well as convergence rates.}
    \label{fig:coconts_multinomial}
\end{figure}

\xhdr{Effect of $r$} A crucial hyperparameter in our approximation is $r$. While we see the effects of $r$ on overall model quality in Sec. \ref{sec:ablations}, we study the effect of $r$ in a more controlled way when learning a single 10-class multinomial. By increasing $r$, as shown in Fig. \ref{fig:coconts_multinomial}, we find that both convergence rate as well as KL divergence between learned and actual multinomial consistently improves.

\subsection{Additional Discussion on Pre-enriching the Dataset}
\label{sec:appendix_preenriching}
\begin{figure}[t]
    \centering
    \includegraphics[width=0.5\textwidth]{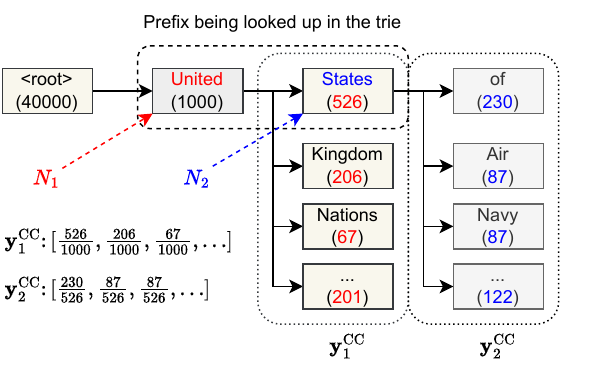}
    \caption{\textbf{Example of trie lookup for prefix ``United States''.} The number in the paranthesis denotes the ``count'' property of the TrieNode as described earlier. Notice the denominator terms of $\yadj{1},\yadj{2}$ carefully. These will be truncated to only include Top-$r$ values.}
    \label{fig:trielookup}
\end{figure}
In this section, we provide a step-by-step walkthrough of Pre-enriching process for the dataset.

First, we assume that the prefix-trie is already created and available in-memory for $k=2$. This trie has a crucial property: the ``count'' attribute (64 bit unsigned integer in our implementation) of each node $N_i$ indicates how many times the prefix, which corresponds to the path from the root to $N_i$, appears in the training dataset $\trainset$. The root node stores the count of total prefixes in the dataset. Figure \ref{fig:trielookup} shows an example trie, where the node associated with the token ``United'' has a count of 1000, while the root node has a count of 40000. This means that there are 1000 sequences in total that begin with the word ``United''. Additionally, the child node ``States'' has a count of 526, indicating that there are 526 sequences that start with ``United States''.

Let's assume that a block (length $L$) beginning with "United States of America $\dots$'' is selected in the pre-enriching second pass over the dataset. 
Since $k=2$, we retrieve from the prefix trie with prefix ``United States'' as shown in Figure \ref{fig:trielookup}. At every level (i.e. highlighted $N_i$), first $\ymulti{i}$ will be created in floating point representation with the same bitwidth as token IDs. In our case, this was fp16. Then, we sort the distribution to get top $r$ token IDs and top $r$ probability values. Storing these on disk will require space equivalent to $2kr$ more tokens. 

After a block of $L$ tokens is read from the input file, it is immediately written as is to the output file. Then we find top $r$ token IDs and probabilities for each $k$ and sequentially write these values to the output file. After all $k$ such distributions are written, we would have written equivalent of $L+2kr$ tokens to the output file.

\subsection{Additional Discussion on Minibatch building with $\yadj{i}$}
\label{sec:appendix_minibatch}

Since we know that the maximum support of all the $\yadj{i}$ distributions is $r+1$, we can ideally easily pass them as key-value pairs to the training loop and calculate KL divergence more efficiently. This causes only $O(kr)$ increase in memory footprint of a batch. However, this can be slightly inefficient since we need to run ``\texttt{gather}'' operation to obtain correct components of the predicted distribution $P_\theta(y_i)$ based on token indices. ``\texttt{gather}'' operation is often slow\footnote{\tiny{\url{https://github.com/pytorch/xla/issues/898}, \url{https://github.com/pytorch/xla/issues/3587}}} on TPU/XLA devices which rely on a predictable dataflow in order to optimize their compute graph. Prior works such as BigBird \citep{zaheer2020bigbird} have resorted to special resizing and converted the operation to continuous memory access. 

Such tricks are harder to implement here without ascertaining an upper bound on the maximum token ID in the support of $\yadj{i}$. Ideally, obtaining such bounds may be useful and possible since the tokenizer (such as WordPiece or BPE) are expected  to assign lower token IDs (earlier ``merges'') to frequent tokens anyway. In our implementation, we initialize a $k\times|\vocab|$ size zero vector and use the ``\texttt{scatter}'' \footnote{\tiny{\url{https://pytorch.org/docs/stable/generated/torch.scatter.html}}} operation to populate counts at correct places. While the  ``\texttt{scatter}'' operation is also slow, we perform it during batch creation on CPU which is latency optimized as opposed to previous proposal which was doing ``\texttt{gather}'' operation on accelerators which are throughput optimized. While this increases the memory footprint of the batch by $O(k|\vocab|)$, we found that using such dense vectors for KL divergence resulted in the model running slightly faster on both TPUs and GPUs.

Since we pack the $[\yadj{1},\dots,\yadj{k}]$ directly into the batch as $|\vocab|$-dimensional vectors, the memory used by the labels in the \sysname\ objective is considerably higher than in the baseline. Despite this, we occupy only 0.5\% of the total TPU/GPU RAM used by the trainer. The baseline (\NT) method takes only $128\times 256\times 2 = 64\text{KB}$ of memory to store labels for a sequence length of 256 and a batch size of 128, assuming
16-bit integer token IDs. In the \sysname\ objective, we provide additional $k$ distributions. Assuming each number in the distribution is a 16-bit float, and considering $k=8$ and GPT2's vocabulary size of $50257$, we occupy approximately $128 \times 8 \times 50257 \times 2 = 98.16\text{MB}$ of additional GPU/TPU RAM per batch during training. On our GPUs, the \NT\ objective utilizes around 72GB of RAM out of the total available 80GB, leaving more than enough room to accommodate all the $k=8$ extra distributions per sequence per batch.

\end{document}